# A multimodal dataset of photoplethysmography and continuous behavioral responses to ASMR and nature videos

**Tushar Das[1], Daigo Hozaki[2], Koushlendra Kumar Singh[1], and Hirohito M. Kondo[2]**

[1]Machine Vision & Intelligence Lab, National Institute of Technology Jamshedpur, Jamshedpur 831014, India

[2]School of Psychology, Chukyo University, Nagoya, Aichi 466-8666, Japan

[*]Corresponding Author: Tushar Das, e-mail: tdas2663@gmail.com

## Abstract

Autonomous Sensory Meridian Response (ASMR) is a somatosensory phenomenon characterized by pleasant tingling sensations and cardiovascular slowing. However, ASMR research has been hindered by a dearth of standardized, open-access multimodal datasets. To address this limitation, we present REST-ASMR (Response to Environmental & Sensory Triggers), a synchronized multimodal dataset designed to capture behavioral reports and physiological dynamics during ASMR, with nature-relaxation videos as control stimuli. The dataset includes high-resolution photoplethysmography (PPG), time-aligned audiovisual stimuli, and continuous subjective annotations from 34 participants. Technical validation showed high stimulus efficacy (97% responder rate), significant stimulus-specific inter-subject agreement ($p < 0.05$), and a robust PPG-derived ASMR-specific cardiovascular deceleration. Additionally, a Bidirectional Long-Short Term Memory model successfully predicted subjective ASMR tingle states, achieving video-level ASMR vs. Nature classification with perfect accuracy and a frame-level global mean accuracy of 75.51%, macro F1-score of 71.86%, and 100% Nature-baseline specificity, under a strict, leakage-free subject-video double-independent 4-fold cross-validation. REST-ASMR constitutes a dense temporal foundation for affective computing, multimodal research, and the development of personalized models of relaxation-related responses.

## Background & Summary

Autonomous Sensory Meridian Response (ASMR) refers to a sensory experience that features distinct tingling sensations, often described as static-like, that start on the scalp and spread along the spine[1]. This experience is often described as a state of "low-grade euphoria," which combines positive feelings with these localized tactile sensations. The term was first used in 2010 to give a clear name to the sensation, although research into its biological basis has only recently gained traction[1]. ASMR is usually triggered by specific audiovisual stimuli like whispering, crisp sounds (such as tapping or crinkling), and experiences that simulate personal attention or grooming[2,3].

While early descriptions of ASMR were mostly based on personal accounts, recent studies have demonstrated its biological foundation. Research monitoring physiological responses has revealed that the ASMR state is associated with specific bodily changes, such as a slower heart rate and increased skin conductance, setting it apart from general relaxation[4]. Neuroimaging studies of individuals sensitive to ASMR have found reduced connectivity in the brain's default mode network (DMN) and increased activity in regions associated with reward and social understanding, such as the nucleus accumbens and the medial prefrontal cortex[5]. Recent research suggests multiple possible benefits of ASMR, including mood improvement[6], increased focus[7], anxiety reduction[8], lower blood[9] pressure, and better sleep[9,10], among others

Furthermore, research has consistently shown natural stimuli to have a relaxing effect on people[11–13]; however, its suggested theoretical underpinnings differ from those of ASMR. While natural environments provide an escape from social situations, ASMR-induced relaxation apparently benefits from socially simulated triggers[4,14]. This contrastive distinction between the two relaxation-inducing phenomena invites research into the mechanisms of both.

Despite these findings, the exact timing and features that lead from an audio-visual trigger to a physiological response remain poorly understood. A significant issue in current affective computing research is the lack of standardized, high-quality, publicly available multimodal datasets containing both ASMR and non-ASMR stimuli. The dearth of such resources has hindered the detailed modelling of ASMR response using statistical, machine learning, or deep learning methods, contrasted against non-ASMR controls, as researchers were often restricted to private datasets or self-reported data[9], limiting the ability to naturally learn the complex and non-linear relationships in sensory experiences. This lack of a publicly available multimodal ASMR dataset hinders the pace of interdisciplinary research required to understand the psychological, physiological, and computational aspects and prospects of the ASMR phenomenon, and to distinguish it from non-ASMR routes of relaxation[15].

To address this deficiency, this paper introduces a validated multimodal dataset, titled Response to Environmental & Sensory Triggers-ASMR (REST-ASMR). This dataset comprises synchronized video, audio, continuous tingle and pleasantness ratings, and photoplethysmography (PPG) recordings collected from 34 healthy participants. To the best of our knowledge, REST-ASMR is the first open-access resource to combine these modalities within a comparative experimental design that includes both Nature (non-ASMR) and ASMR stimuli.

The REST-ASMR dataset serves as a foundational benchmark for interdisciplinary research. It aids in discovering identifiable acoustic patterns and in developing multimodal fusion architectures to examine how visual and auditory cues interact. By integrating simultaneous audiovisual stimuli with high-resolution PPG signals and continuous subjective ratings, this dataset enables the study of the causal dynamics of sensory-induced relaxation over time and the identification of predictive physiological biomarkers. While the audiovisual stimuli serve as the primary ASMR-eliciting triggers, the PPG modality provides invaluable cardiovascular insights into the phenomenon[16,17]. It supports the development of personalized audio-medicine[7,18] and digital therapeutics[10] for stress and insomnia, while providing a framework for studying the differences between environment-led tranquillity and ASMR-induced relaxation. Additionally, the dataset's high temporal precision opens doors for innovations in generative deep learning[19]. This includes creating physiology-guided ASMR content[20,21] and designing adaptive systems that use real-time physiological feedback to optimize heart rate reduction[22] through tailored acoustic modulation. This resource facilitates the advancement of cross-disciplinary studies in affective computing, sensory neuroscience, and human-computer interaction[23–25].

# Methods

## Participants

The study cohort consisted of college students recruited from the university population. To ensure statistical robustness, a priori power calculations were performed using G*Power (v3.1.9.2)[26], which indicated that a sample size of 34 would achieve a power of 0.80 ($\alpha = 0.05$) for detecting medium-sized effects (Cohen's $f = 0.25$) in a repeated-measures design. From an initial pool of 35 recruits, data from one individual were discarded post-acquisition on account of physiological irregularities (resting pulse rate consistently exceeding 110 bpm). The final validated dataset comprises 34 participants (self-reported gender: 18 men, 16 women; mean age = 25.2 ± 4.6 years; range: 19–35 years). Eligibility criteria required participants to be right-handed and possess normal hearing and normal or corrected-to-normal vision. To prevent familiarity bias, potential participants were screened to ensure they had not previously viewed the specific environmental stimuli selected for the control condition.

## Ethics Statement

All experimental protocols were reviewed and approved by the Research Ethics Committee of Chukyo University (Approval No. RS21–026). In accordance with the Declaration of Helsinki, written informed consent was obtained from all participants, who received monetary compensation (JPY 1,000) for their participation.

## Stimuli Selection

The audiovisual stimuli consisted of eight distinct 60-s clips, stratified into two experimental categories:

1. **ASMR Triggers (N=4):** Four video clips (labeled ASMR1 to ASMR4) were derived from open-access content hosted on YouTube, specifically videos distributed under Creative Commons licensing (https://www.youtube.com/watch?v=7MZtaAgqoTY and https://www.youtube.com/watch?v=asGLp12NSIE). The specific excerpts were curated based on their efficacy in eliciting tingling sensations, as verified by three of the authors. The selected clips featured non-vocal, binaural acoustic triggers, predominantly brushing and tapping sounds. To isolate the acoustic-tactile trigger mechanism and eliminate social confounding variables, the visual component was cropped to exclude the performer's facial features.

2. **Environmental Controls (N=4):** To provide a non-social relaxation baseline, four nature sequences (labeled nature1–nature4) were sourced from YouTube. These clips depict abiotic environments, specifically a flowing brook, ocean waves, a bonfire, and a wind-swept landscape, devoid of human presence[27].

## Experimental Procedure

The data acquisition session lasted approximately 12 minutes and utilized a counterbalanced block design. The procedure included two main 4-minute stimulus blocks separated by resting intervals, structured as: Baseline Rest (60s) → Block A (240s) → Washout Rest (120s) → Block B (240s) → Final Rest (60s).

To mitigate order effects, participants were alternately assigned to one of two presentation sequences:

- **Sequence 1**: ASMR stimuli presented in the first block, followed by Nature stimuli.
- **Sequence 2**: Nature stimuli presented in the first block, followed by ASMR stimuli.

Within each block, the four category-specific videos were presented in a pseudo-randomized order. Participants were given a visual indication prior to each test session that specified whether ASMR or nature videos would be shown. The complete 12-minute session timeline and counterbalanced block design are illustrated in **Figure 1**.

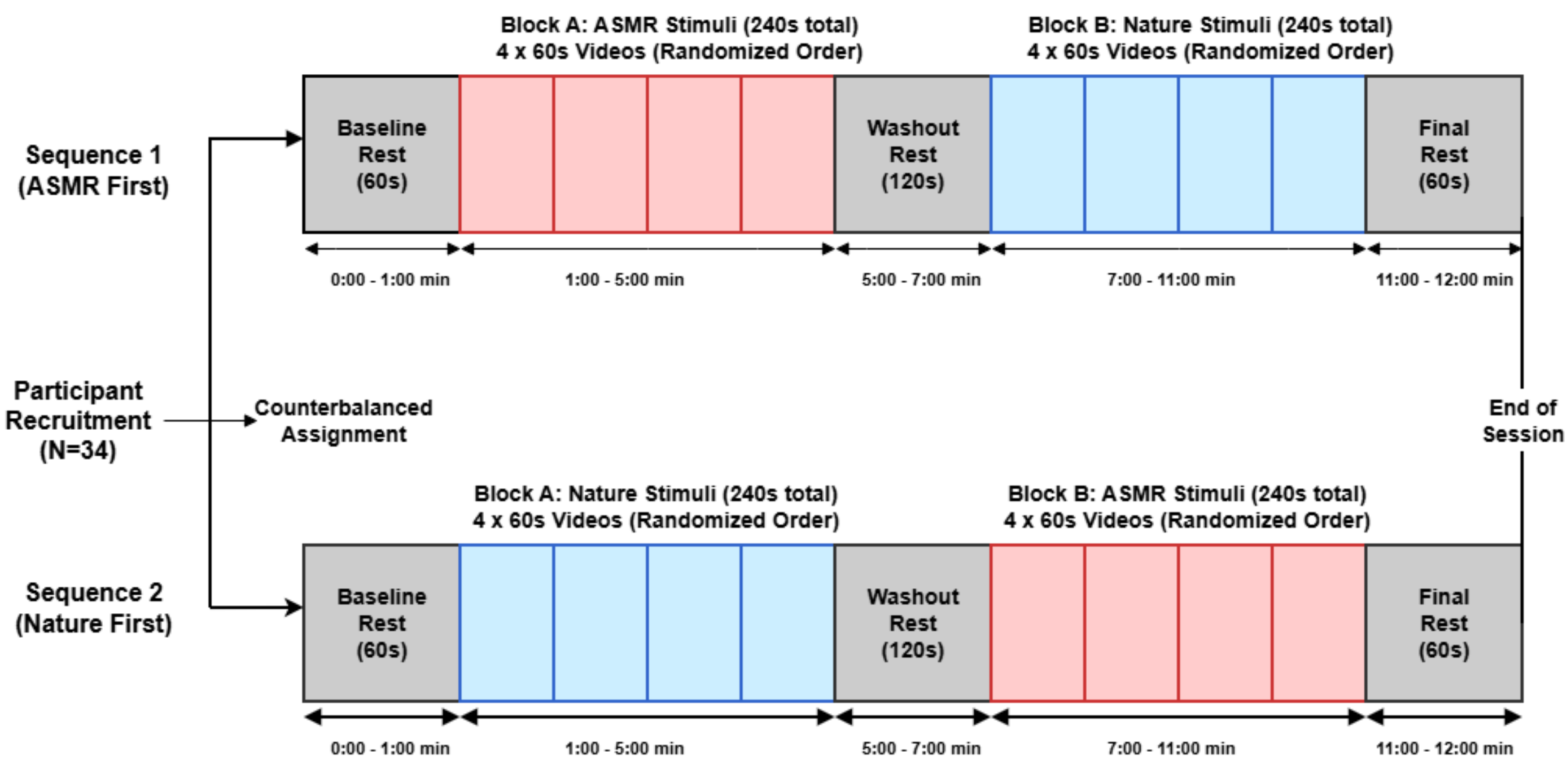

***Figure 1***. *Experimental protocol timeline and counterbalanced block design. The 12-minute data acquisition session utilized a repeated-measures design to alternate the presentation order of the stimuli. To mitigate potential order and carryover effects, the 34 participants were counterbalanced into two groups. Sequence 1 (top) presented the ASMR stimuli block first, while Sequence 2 (bottom) presented the non-social Nature control stimuli block first. Each session commenced with a 60-second*

*baseline resting period, followed by the first 240-second stimulus block (comprising four category-specific 60-second videos presented in a pseudo-randomized order). A 120-second washout resting period separated the two stimulus blocks to allow cardiovascular metrics to return to baseline. Following the presentation of the second 240-second stimulus block, a final 60-second resting period concluded the acquisition session.*

During resting intervals, a blank screen was displayed to facilitate physiological recovery. Stimulus delivery and timestamp logging were managed via Presentation software (Neurobehavioral Systems, Berkeley, CA, USA), ensuring precise synchronization.

Visual stimuli were rendered on an LCD monitor with a temporal resolution of 60 Hz and a spatial resolution of 1280 × 720 pixels. Participants were seated at a fixed viewing distance of approximately 57 cm, resulting in a calculated visual angle of 16.0° × 9.0°. To maintain the spatial immersion of the binaural recordings, audio stimuli were delivered dichotically via Sennheiser HD 599 open-back headphones.

Note: The raw physiological and behavioral data described in this dataset were originally acquired during the experimental procedures of a prior related study[14]; however, the synchronized, multimodal dataset, the associated preprocessing codebase, and the complete open-access repository are being published and technically validated in this paper for the first time. Also, during the preparation of this manuscript, the authors used AI-assisted tools, Grammarly 1.2.220 and Google Gemini 3 Pro, for grammar and language refinement, formatting, and clarity. All content was created, reviewed, and edited by the authors, who take full responsibility for the final manuscript.

# Data Acquisition

## Behavioral Data

During stimulus playback, participants registered their subjective experience in real-time using a discrete 4-point Likert scale (1: “none,” 2: “slightly,” 3: “moderate,” and 4: “very”). Participants rated "Tingling Intensity" (for ASMR videos) and "Pleasantness" (for Nature videos) using their right fingers on a standard keyboard at a 1000 Hz sampling rate, with keystrokes logged as time-stamped events. At the start of the experiment, participants engaged in practice keystrokes to familiarize themselves with the continuous assessment interface and were instructed to minimize motor activity during the data collection phase. It is worth noting that while the log files record these ratings on a scale of 1 to 4, as detailed in the data processing section, these ratings were linearly mapped to the 0-to-3 range before subsequent technical validation.

## Photoplethysmography Data (PPG)

Cardiovascular data were acquired using a Biopac Systems MP36 acquisition unit (Goleta, CA, USA). An SS4LA PPG transducer was secured to the distal phalanx of the participant's left middle finger to minimize motion artifacts from the dominant right hand used for behavioral inputs. The raw blood volume pulse was digitized at a sampling rate of 2000 Hz. While online filters (0.5–35 Hz) were available for signal monitoring, the raw unfiltered data were retained for the dataset.

# Data Processing and Temporal Alignment

To convert the raw event logs and physiological recordings into a synchronized dataset ready for statistical and machine-learning applications, a multi-stage preprocessing pipeline was implemented using Python. This pipeline addressed three key challenges: variable video initialization latency, sparse behavioral sampling, and cross-modal synchronization.

## Behavioral Data Alignment (Log Parsing)

Raw event logs were parsed to separate stimulus-onset and participant-response events. A systematic Index-Based Synchronization algorithm was applied to correct for variable buffering delays observed during the system initialization phase. The key steps are outlined below.

1. **Latency Correction**: A system initialization delay of ~4.3 s caused the first video trial to begin late relative to the fixed experimental schedule. The next trial was scheduled to start at an exact time, so the software ended the first video early to keep the timing aligned across trials. To prevent temporal misalignment and data leakage into the following trial, the valid duration of the first stimulus (Sequence Index 0) for every participant was clamped to 55.7 seconds. Feature extraction was restricted to this window, truncating the final 4.3 s of the video file. All subsequent videos (Indices 1-7) retained the full 60.0-second duration for each participant.

2. **State-Aware Buffering**: To account for physiological settling time, the first 5.0 seconds of each clip were designated as a warm-up buffer. This buffer period, instead of being discarded, was scanned for participant interaction. The final keypress recorded within this 5-second window was carried forward to establish the initial intensity state (t = 0) of the valid data epoch. If no interaction occurred, the initial state defaulted to 0 (None). As a result, the final analyzed epoch for the initialization trial (the first video of each participant) spanned from t = 5.0 s to t = 55.7 s (yielding 50.7 s of valid data), while the continuous trials (the remaining 7 videos of each participant) spanned from t = 5.0 s to t = 60.0 s (yielding 55.0 s of valid data).

3. **Dense Upsampling**: To enable frame-by-frame analysis, the sparse, asynchronous keypress events were upsampled to a dense 10 Hz time-series (0.1 second steps). A zero-order hold logic was applied, where the label for any given timestamp t was maintained at the value of the most recent past keypress until a new input was registered.

## Physiological Signal Processing

Biopac Student Lab Pro (v4.1) was used to export raw PPG waveforms as text files. To guarantee precise synchronization with the updated behavioral timelines, PPG data were processed using a Timeline Lookup technique. Each Subject ID and Video ID was linked to the exact start and end timestamps (with a precision of 0.1 ms) obtained from the latency-corrected logs using a lookup map. The Biopac recordings' Channel 2 (CH2) was used by the preprocessing pipeline. The continuous PPG signal in this channel has been pre-filtered with a 0.5–35 Hz bandpass filter to attenuate high-frequency noise and baseline drift. These absolute windows were used to slice the 2,000 Hz PPG signal, ensuring that the biological data were phase-locked to the behavioral labels. To match the temporal resolution of the behavioral and audiovisual features, the PPG slices were downsampled from 2,000 Hz to 10 Hz using Fourier method resampling (via scipy.signal)[28]. Finally, to mitigate inter-participant variability in baseline pulse amplitude, Z-score normalization was applied to each individual clip, scaling the signal to have a mean of 0 and a standard deviation of 1. This video-level PPG normalization mitigated data leakage, as the technical validation is performed at the participant and video level. The final processed dataset consists of aligned time-series matrices where every row represents a 100 ms snapshot containing the Subject ID, Video ID, Normalized PPG value, and the instantaneous Subjective Intensity Label. The complete parallel processing and temporal synchronization pipeline for all three data modalities is summarized in **Figure 2**.

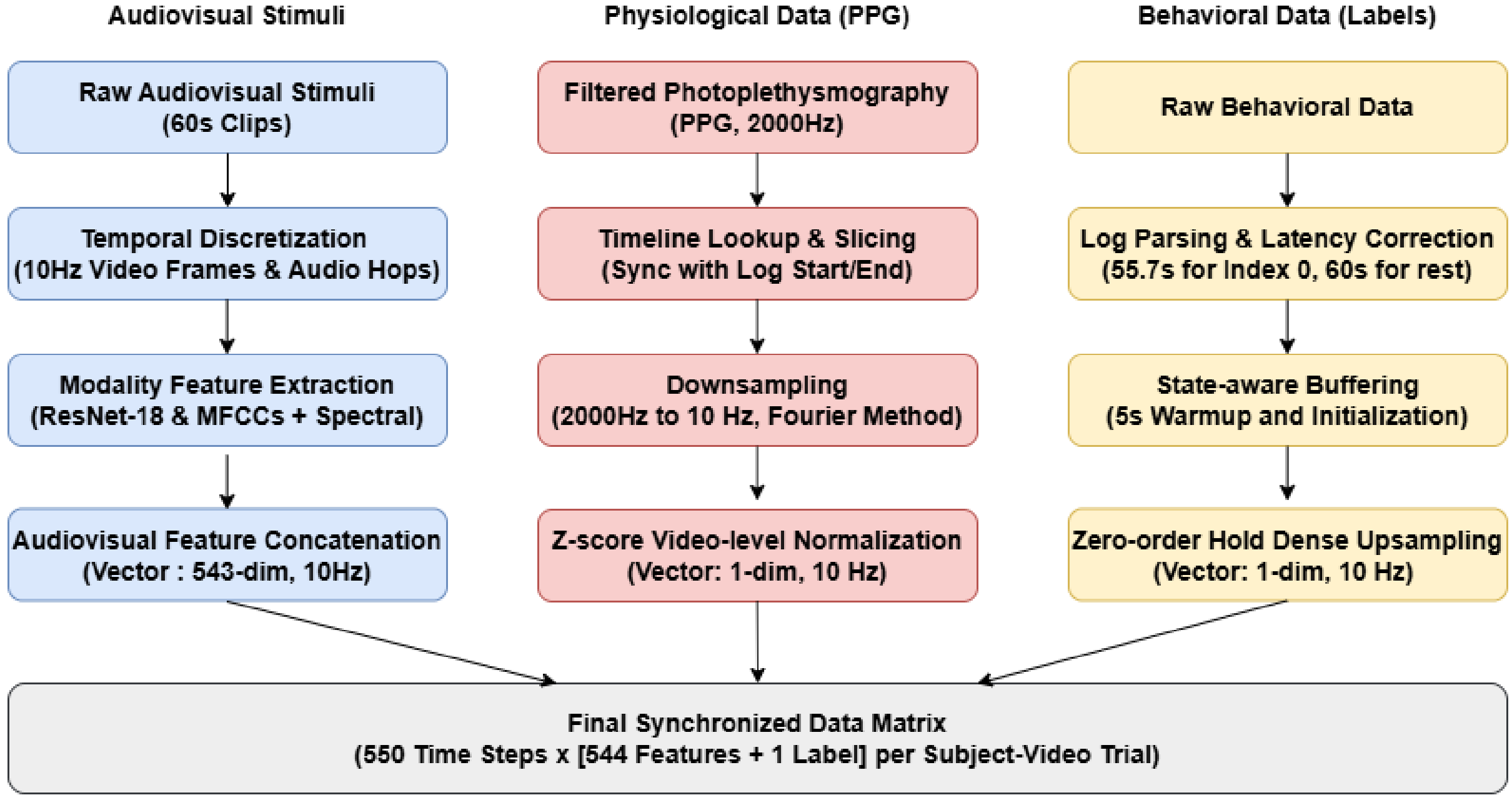

***Figure 2.*** *Multimodal data processing and temporal alignment pipeline. The parallel processing streams used to synchronize various data modalities into a single, dense matrix at a consistent 10 Hz temporal resolution are depicted in the schematic. In order to extract spatial (ResNet-18) and acoustic (MFCCs and spectral) features, the audiovisual stream (left) discretizes raw stimuli. The physiological stream (center) uses video-level Z-score normalization and Fourier-based downsampling to separate the raw 2,000 Hz photoplethysmography (PPG) signal. The behavioral stream (right) creates a dense continuous label by parsing sparse subjective rating logs, applying the required latency corrections, enforcing a 5-second warm-up buffer, and using zero-order hold upsampling. A final standardized data matrix with 550 time steps by 545 dimensions (544 predictive features and 1 binary state label) is created by the convergence of these three independent streams for each participant-video trial.*

# Data Records

The REST-ASMR dataset generated and analysed during the current study is available in the Zenodo repository[29] at https://doi.org/10.5281/zenodo.18881334.

Note: While individuals are generally referred to as 'participants' throughout the behavioral and physiological descriptions in this manuscript, the term 'subject' is retained within the machine learning validation sections (e.g., 'test subjects') to align with standard

computational terminology. Also, the raw data files, event logs, and the accompanying Python codebase utilize the variable 'Subject ID' (e.g., 001 to 035) to maintain programmatic consistency.

The dataset is archived in a standard hierarchical structure containing three primary directories: log, ppg, and stim.

1. **Log Files (/log):** This directory houses 34 event logs generated by the Presentation software. Files follow the naming convention: [SubjectID]-[Condition]-[Date]-[Hash].log.

    - **SubjectID**: A unique identifier ranging from 001 to 035, excluding 002.
    - **Condition:** Denotes the block order with odd Condition IDs corresponding to ASMR-first order (ASMR block followed by nature block) and even IDs corresponding to Nature-first order.
    - **Structure:** Each log is a tab-separated text file containing multiple columns. The important ones are:
        - Time: Event timestamp (in 0.1 ms units).
        - Event Type: Categorical markers for Pulse (system heartbeat), Video (stimulus onset), and Response (user input).
        - Code: Specific identifiers for the stimuli for Video events (vid1–vid4 for ASMR; vid5–vid8 for Nature) and behavioral ratings for Response events (1–4).

2. **Physiological Data (/ppg):** This directory contains 34 raw physiological recordings named [SubjectID]_csv.txt. SubjectID ranges from 001 to 035, except 002.

    - **Format**: ASCII text files with a header specifying the source (.acq) and sampling interval (0.5 ms).
    - **Channels:**
        - Column 1: Time elapsed (minutes).
        - Column 2 (CH1): Raw PPG amplitude (mV).
        - Column 3 (CH2): Pre-filtered PPG amplitude (mV) (0.5–35 Hz bandpass).
    - **Alignment**: The start of the file (t=0) is synchronized with the experiment start time recorded in the corresponding log file.

3. **Stimuli (/stim):** This directory contains the source media files in .avi format.

    - ASMR: ASMR1.avi to ASMR4.avi.
    - Nature: nature1.avi to nature4.avi.

# Technical Validation

This dataset's main goal is rooted in the need to build a multimodal foundation that will enable the modeling and analysis of the intricate, somatic neurophysiological dynamics of the ASMR phenomenon. A tripartite technical validation protocol was used to guarantee the dataset's validity, reliability, and algorithmic utility. To verify that the chosen stimuli successfully and reliably elicited the desired sensations, behavioral efficacy and inter-subject agreement were first assessed. Second, the continuous photoplethysmography (PPG) streams were evaluated to verify sensor integrity, confirming the parasympathetic cardiovascular deceleration that is uniquely co-constituted with the ASMR state. Lastly, baseline machine learning architectures were trained to assess the multimodal features' predictive utility, demonstrating that the synchronized physiological and audiovisual streams contain useful representations for sophisticated sequence-to-sequence modeling.

## Stimulus Efficacy and Behavioral Agreement

A behavioral efficacy analysis was carried out to confirm both the epistemic reliability of the subjective annotations and the affective effectiveness of the chosen audiovisual stimuli. Establishing a high baseline responder rate is essential for the dataset's viability because the ASMR sensation is notoriously vulnerable to environmental suppression in normative laboratory settings. A near-perfect elicitation rate was found at the participant level: 33 out of 34 participants, or 97.06% of the cohort, successfully experienced and actively reported the targeted tingling sensation during at least one ASMR trial. Similarly, during the Nature control trials, 97.06% of participants actively reported generalized visual pleasantness. This high degree of agreement suggests ongoing participation, demonstrating that the physiological baselines do not reflect laboratory-induced fatigue but rather attentive participants actively using the rating system.

At the stimulus level, the dataset demonstrates high signal content. The most potent ASMR stimulus (Video 2) triggered active tingles in 94.12 percent of the cohort. Similarly, the highest-performing Nature control stimulus (Video 6) received active pleasantness ratings from 97.06 percent of the cohort.

Importantly, when the targeted sensations were experienced, they produced ample data for the training algorithms. Once triggered, the ASMR tingling sensation lasted an average of 38.54 ± 15.57 s per 55.0-s trial, while the Nature pleasantness was reported for a mean duration of 52.12 ± 8.60 s. This sustained period of active subjective reporting provides evidence that the dataset is highly dense in positive-class physiological inscriptions. Overall, these strong behavioral metrics confirm that the experimental protocol effectively overcame laboratory limitations, resulting in a high-quality dataset suited for training data-intensive machine learning models.

To further validate the temporal consistency of the subjective annotations, a Leave-One-Out Inter-Subject Agreement (LOO-ISA) analysis was enacted on the continuous behavioral traces. This analysis interrogates whether the reported ASMR sensations were simply generalized affective states or precise, stimulus-locked physiological reactions. For each ASMR stimulus, a single participant's continuous rating array was isolated and correlated (using Pearson's r) against the averaged rating array of all other participants who viewed the same video. This leave-one-out process was repeated for every participant, and the resulting correlation coefficients were Fisher Z-transformed to generate a group-level distribution of agreement. To ensure strict statistical rigor without assuming a normal distribution, a 10,000-iteration bootstrap resampling test was applied to these coefficients to ascertain if the group agreement was significantly greater than zero.

The analysis revealed a statistically significant temporal alignment across all targeted ASMR stimuli ($p < 0.05$). The alignment was exceptionally strong for the primary elicitation stimuli, most notably in Video 3 (Mean $r = 0.539$, 95% confidence interval (CI) [0.377, 0.654], $p < 0.001$) and Video 2 (Mean $r = 0.424$, $p < 0.001$), as visualized in the group-level mean time-series (**Figure 3**). This significant inter-subject agreement confirms that the continuous subjective labels closely match specific audiovisual events. Therefore, the behavioral arrays serve as reliable markers for high-resolution temporal machine learning models.

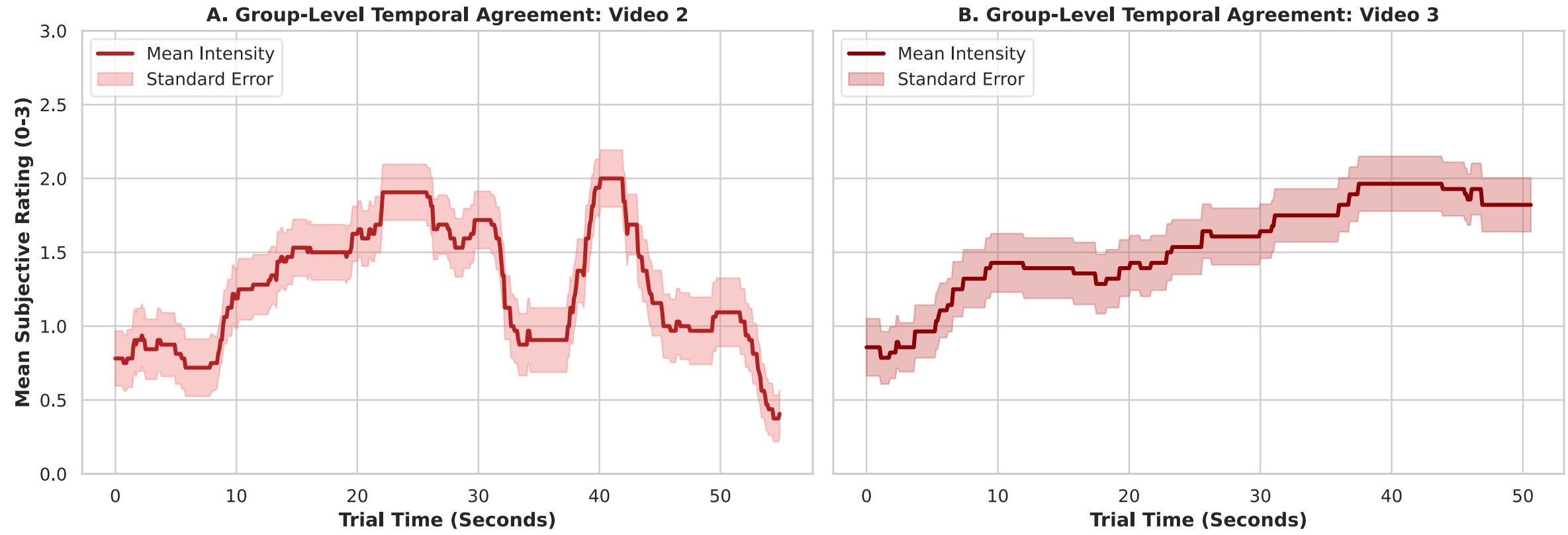

*Figure 3. Group-level temporal dynamics of continuous behavioral annotations. Mean time-series plots for the two highest-performing ASMR stimuli, (A) Video 2 and (B) Video 3, visualize the temporal consensus across all active participants. The shaded regions represent the standard error of the mean (SEM). The contrasting temporal architectures, ranging from multi-phasic, episodic triggering in Video 2 to steady, accumulative physiological build-up in Video 3, highlight the diverse temporal dynamics captured within the dataset, together with synchronized Inter-Subject Agreement across different trigger types.*

## Physiological Signal Integrity and Cardiovascular Response

Prior to downstream algorithmic modeling and the subsequent mapping of behavioral correlates, the raw physiological streams were evaluated to establish structural integrity. Analyzing the complete set of 2000 Hz PPG recordings from all trials (29,627,861 total frames) showed an impressive sensor continuity rate of 99.96 percent. This metric, computed by analyzing the first derivative of the raw waveforms to detect sensor disconnects, voltage clipping, or motion-induced flatlines, indicates that signal loss was negligible throughout the experimental protocols, thereby providing a reliable basis for cardiovascular analysis.

To confirm the biological effects of the stimuli and ensure the dataset recorded a distinct physiological state linked to the ASMR experience, a short-term cardiovascular correlation analysis was performed. A 5.0-second sliding window with a 1.0-second step stride was applied to the continuous PPG streams. Within each window, a 0.5 to 4.0 Hz bandpass Butterworth filter was applied to isolate the pulsatile cardiovascular frequency. Systolic peaks were detected to extract valid peak-to-peak (pulse-to-pulse) intervals, from which the instantaneous Pulse Rate (PR) was computed. Concurrently, the mean subjective ordinal rating (ranging from 0 to 3) was aggregated for each corresponding 5.0-second window.

To quantify the relationship between stimulus intensity and the ensuing cardiovascular state, the Spearman rank correlation coefficient (r) between the instantaneous PR and the ordinal subjective ratings was calculated independently for each participant across both the ASMR and Nature control conditions. Spearman correlation was specifically utilized to account for the non-parametric, ordinal nature of the behavioral labels. Importantly, to forestall the artificial inflation of degrees of freedom induced by the autocorrelation of overlapping sliding windows, statistical testing was strictly constrained to the participant level. To evaluate the group-level divergence between the two conditions, the participant-level correlation coefficients were converted to a normal distribution utilizing Fisher's Z-transformation ($z = \text{arctanh}(r)$).

A paired samples t-test conducted on the Z-transformed coefficients indicated a statistically significant difference in the cardiovascular response dynamics between the two conditions ($t = -2.085$, $p = 0.0454$). Because this analysis interrogated a single, a priori physiological hypothesis regarding PR divergence, multiple comparison corrections were rendered unnecessary. The mean PR correlation during the ASMR condition was negative ($r = -0.0921$), showing that as the subjective intensity of the ASMR tingle increased, participants' pulse rates decreased. In contrast, the Nature control condition exhibited a slight positive correlation ($r = 0.0556$). This significant divergence, as visualized in **Figure 4**, confirms that the dataset successfully captured the sedative parasympathetic cardiovascular deceleration uniquely associated with the ASMR phenomenon documented in prior literature[14], demonstrating that the physiological features hold strong, biologically relevant signals that differ from general positive feelings or visual stimulation

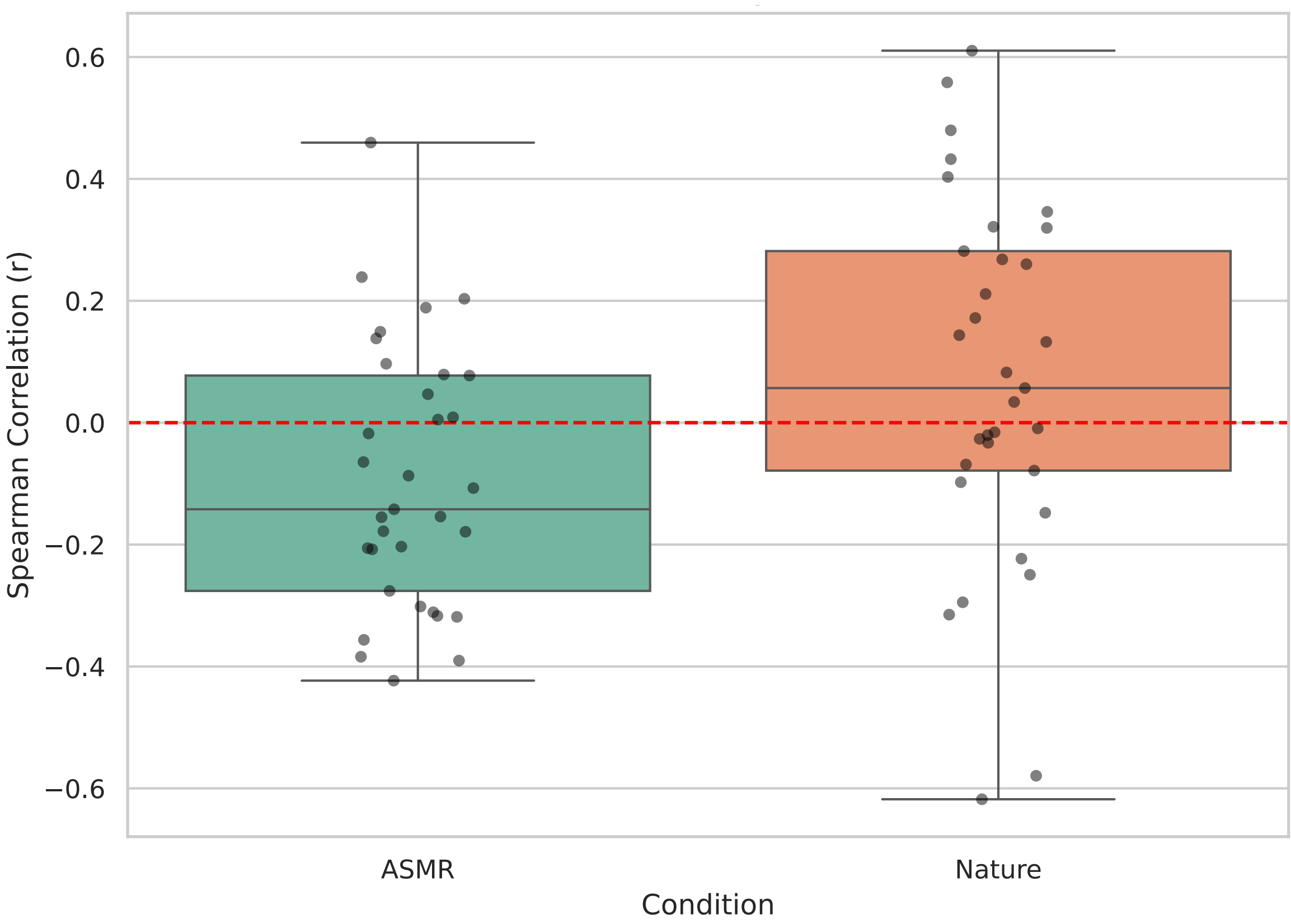

***Figure 4***. *Participant-level cardiovascular correlation across experimental conditions. Boxplots illustrating the distribution of Spearman rank correlation coefficients (r) between instantaneous pulse rate (PR) and subjective intensity ratings for the ASMR and Nature control stimuli. The ASMR condition exhibits a negative mean correlation, reflecting the parasympathetic cardiovascular deceleration accompanying the ASMR tingle. Conversely, the Nature condition trends positively, consistent with mild visual engagement and baseline physiological arousal. Individual participant data points are overlaid to demonstrate the population variance. The group-level difference is statistically significant (p = 0.0454) based on a paired t-test of the Fisher Z-transformed coefficients.*

## Baseline ML Predictive Utility and Modality Ablation Experiments

A multimodal feature-extraction pipeline was implemented to compress high-dimensional raw audiovisual and physiological streams into a dense, learnable representation. For the visual aspect, video frames were sampled at 10 Hz, resized to 224x224 pixels, and normalized using standard ImageNet statistics. Spatial features were extracted using a pre-trained ResNet-18[30] convolutional neural network with the final classification layer removed, yielding a 512-dimensional vector per frame. For the acoustic modality, audio tracks were processed using the *librosa* library[31] with a 10 Hz hop length to ensure they stayed in sync with the visual frames. Each time step generated a 31-dimensional acoustic feature vector, which included 20 Mel-Frequency Cepstral Coefficients (MFCCs), root-mean-square energy, spectral centroid, spectral bandwidth, zero-crossing rate, and spectral contrast bands. Finally, the Z-score normalized continuous PPG signal provided a 1-dimensional physiological feature. These streams were concatenated horizontally to produce a unified 544-dimensional multimodal feature vector per 100-millisecond time step.

To ensure uniform tensor dimensions across all 272 experimental trials (34 participants observing 8 videos each), the extracted sequences were either padded or truncated to a fixed length of 550 time steps, or 55.0 seconds. Furthermore, since a primary objective of this dataset is to enable sequence-to-sequence ASMR modelling and the exploration of the biological underpinnings of ASMR, characterized by the tingling sensation, this technical validation focused on formulating a frame-by-frame binary prediction task. Hence, for each ASMR video, binary labels were generated by thresholding the subjective participant ratings (rating 0: no tingle; ratings 1, 2, and 3: tingle); all nature videos were labelled as "no tingle" to establish a non-triggering baseline.

This procedure led to a final standardized data matrix with 550 rows and 544 columns for each trial. It included a sequence of binary labels that indicate whether a participant self-reported the presence (1) or absence (0) of an ASMR tingle. Random seed (42) was used throughout to ensure reproducibility. The complete multimodal feature extraction pipeline and subsequent predictive architecture are illustrated in **Figure 5**.

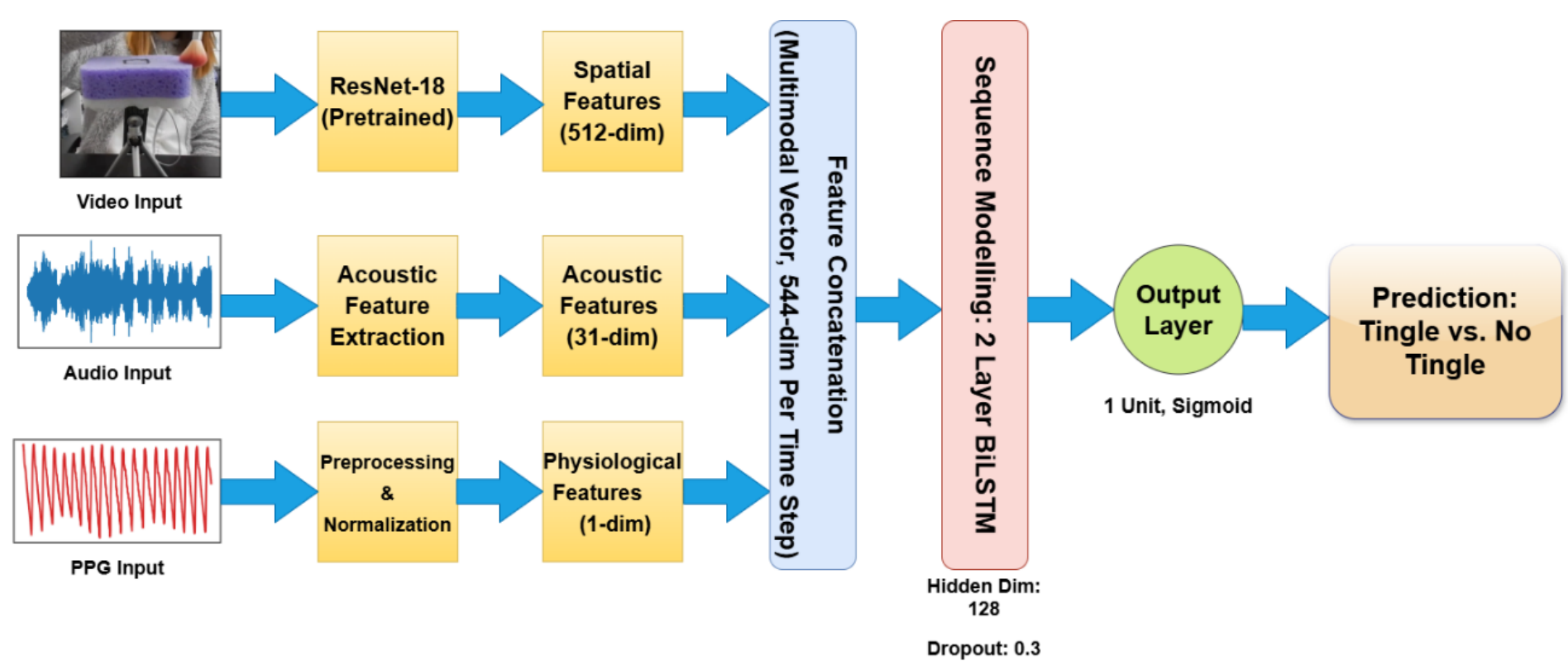

***Figure 5**. Multimodal deep learning architecture for frame-by-frame ASMR state prediction. Raw audiovisual and physiological streams are synchronized and sampled at 10 Hz. For the visual modality (top), a pre-trained ResNet-18 network extracts 512-dimensional spatial features from individual video frames. The acoustic modality (middle) extracts a 31-dimensional feature vector comprising Mel-Frequency Cepstral Coefficients (MFCCs) and spectral properties. The physiological modality (bottom) utilizes Z-score normalized continuous photoplethysmography (PPG) signals to yield a 1-dimensional feature. These discrete streams are horizontally concatenated to form a unified 544-dimensional multimodal feature vector for each 100-millisecond time step. A two-layer Bidirectional Long Short-Term Memory (BiLSTM) network, with a hidden dimension of 128 and a dropout rate of 0.3, processes this dense sequential representation to model complex temporal dependencies. A fully connected classification layer with a sigmoid activation function then outputs the frame-level binary prediction, classifying the participant's state as either experiencing a tingle (1) or not (0).*

A sequence-to-sequence deep learning architecture was employed to evaluate the baseline predictive utility of the extracted features. The network consisted of a two-layer Bidirectional Long Short-Term Memory (BiLSTM) module[32] with a hidden state dimension of 128 and a dropout rate of 0.3, followed by a fully connected linear classification head. To address the inherent class imbalance of the subjective tingling responses, a dynamically weighted cross-entropy loss function was utilized.

The network was trained on an NVIDIA T4 GPU and optimized with the Adam[33] optimizer at a learning rate of 0.0005 over 35 epochs, with a batch size of 16.

To prevent data leakage and ensure strong real-world generalization estimates, a strict, double-independent 4-fold cross-validation protocol was executed. In each fold, the test set was isolated by simultaneously withholding 25 percent of the participants and a specific pair of stimuli (one ASMR video and one Nature control video), unique to the particular fold. Folds 1 through 3 had 24 train subjects (participants), 2 validation subjects, and 8 test subjects; fold 4 had 22 train subjects, 2 validation subjects, and 10 test subjects to account for the remaining 2 subjects left out during the division of 34 subjects into four 8-subject test sets. The first ASMR video (vid1) and the first Nature video (vid5) were the test videos for fold 1; the fourth ASMR video (vid4) and the fourth Nature video (vid8) were for fold 4. Videos numbered (i) (ASMR) and (i+4) (Nature) were used as test videos for fold (i). An inner validation split was also performed on the remaining training pool, reserving 20 percent of the training subjects. This inner validation set was used to monitor out-of-sample generalization and to select the optimal model state based on the macro F1-score, precluding overfitting to subject-specific physiological baselines.

To measure population-level uncertainty and rigorously validate the predictive capability of this primary multimodal model while accounting for the high temporal autocorrelation inherent in frame-by-frame continuous data, 95 percent confidence intervals were estimated via non-parametric bootstrap resampling at the participant-level (2,000 resamples). Statistical significance versus chance was assessed utilizing a sequence-level permutation test (5,000 permutations). To create a strong null distribution that reflects the natural temporal dynamics and class imbalance of the responses, entire label sequences were randomly shuffled across trials, rather than permuting individual frames. The effect size of the multimodal network's predictive capability was subsequently quantified using Cohen's d relative to this permutation null distribution.

To further gauge the specific predictive value of the multimodal dataset and to isolate the contributions of individual sensory streams, an ablation study was incorporated into the validation protocol. Two unimodal baseline models were trained: one using only the 512-dimensional spatial features from video and another using the 31-dimensional acoustic feature set. Both unimodal networks followed the same Bidirectional LSTM architecture, hyperparameter configurations, and double-independent 4-fold cross-validation as the complete multimodal fusion model.

Additionally, to showcase the dataset's versatility and establish a non-temporal tabular baseline, an Extreme Gradient Boosting (XGBoost) classifier[34] was implemented. This allows for the quantification of predictive power derived solely from instantaneous physiological and audiovisual states, stripped of temporal sequence context. The 3D sequence tensors were flattened into independent 2D observations, treating each 100-millisecond frame as a discrete data point. The XGBoost model was trained with 100 estimators, a maximum depth of 6, and a learning rate of 0.1. To ensure strict comparability, this non-temporal baseline used the same feature inputs, dynamically calculated class weights to address target imbalance, and was evaluated using the same double-independent 4-fold cross-validation framework as the deep learning models.

Model efficacy across all experimental configurations was evaluated using a suite of performance metrics. Because the frame-by-frame binary classification task is inherently imbalanced (with "no tingle" states naturally outnumbering active "tingle" states), macro-averaged metrics were prioritized to provide an unbiased assessment of predictive performance across both classes. The main evaluation criteria included Global Accuracy, Global Macro F1-score, Global Macro Precision, and Global Macro Recall, aggregated across all four validation folds. Moreover, a specific "Nature Specificity" metric was calculated, defined as the predictive accuracy assessed only on the non-triggering control videos (Nature). This metric was designed to verify that the networks learned true ASMR-triggering patterns rather than defaulting to a spurious positive prediction bias.

While frame-by-frame analysis provides granular temporal validation, ASMR triggers are inherently sparse and episodic; a participant may experience tingling sensations for only a fraction of the stimulus duration. To validate the dataset's utility for macro-level trial classification, a peak-detection strategy was incorporated into the evaluation framework to convert continuous frame-level predictions into discrete video-level binary classification. A complete 55-second video trial was labeled as successfully eliciting ASMR if at least 15 percent of its frames (equivalent to 82.5 frames) were predicted as active tingle states. Importantly, to prevent optimization bias and data leakage, this 15 percent peak-detection threshold was established a priori and was not empirically tuned or optimized using any of the held-out test sets. This video-level classification protocol was applied uniformly across the double-independent 4-fold cross-validation framework (a leave-one-stimulus-pair-out (1 ASMR and 1 Nature video as above) setup, ensuring the network was evaluated on entirely unseen videos). To quantify the uncertainty and statistical significance of this macro-level classification, non-parametric bootstrap resampling (2000 resamples) and permutation testing (5000 permutations) were performed on the video-level predictions. The comparative predictive outcomes of these frame-level and video-level evaluations are detailed in the results section that follows.

## Results

### *Multimodal Baseline Performance and Statistical Significance Results*

The baseline predictive utility of the synchronized multimodal dataset was established using the full-fusion Bidirectional LSTM network, which combines spatial, acoustic, and physiological features. Across the double-independent 4-fold cross-validation protocol, the model achieved a mean global accuracy of 75.51 ± 5.63% and a mean global macro F1-score of 71.86 ± 8.86%. The high mean global macro recall of 81.45 ± 6.34% shows strong sensitivity to the minority tingle class. This outcome confirms that the extracted continuous features have learnable temporal patterns that correspond to subjective behavioral labels.

To ensure these predictive capabilities were not artifacts of temporal autocorrelation or isolated subject variance, rigorous statistical testing was applied to the aggregated predictions. The sequence-level permutation test yielded a p-value of $< 0.0002$ across all evaluation metrics, conclusively rejecting the null hypothesis that the model predictions were driven by chance. Additionally, the effect size, quantified by Cohen's d, was exceptionally large ($d = 5.65$ for the macro F1-score), demonstrating a significant divergence from the random-guessing baseline. The 95 percent confidence interval for the macro F1-score, estimated via participant-level bootstrap resampling, confirming the dataset's stability and generalizability across different participant subsets.

A key validation milestone for the dataset was the Nature Specificity metric, which assessed the model's false-positive rate only on non-triggering control videos. The full fusion model achieved a perfect specificity of 100.00 ± 0.0 % with an absolute 95 percent confidence interval of [100.0, 100.0] and an effect size of d = 5.34. This total rejection of false positives during the control condition clearly shows that the dataset captures real, biologically based ASMR-triggering patterns instead of allowing networks to drift into false positive predictions. The complete statistical analysis results of the full multimodal fusion model are summarized in Table 1.

*Table 1 Advanced Statistical Analysis of the Full Multimodal Fusion Model*

| EVALUATION METRIC (Global) | 4-Fold Mean± SD (%) | 95% CI | P-VALUE | COHEN'S D |
|---|---|---|---|---|
| **Accuracy** | 75.51 ± 5.63 | [69.8, 80.7] | < 0.0002 | 5.65 |
| **Macro F1** | 71.86 ± 8.86 | [65.0, 79.5] | < 0.0002 | 5.65 |
| **Macro Precision** | 73.09 ± 8.42 | [66.9, 79.3] | < 0.0002 | 5.65 |
| **Macro Recall** | 81.45 ± 6.34 | [79.2, 85.5] | < 0.0002 | 5.65 |
| **Nature Specificity** | 100.0 ± 0.0 | [100.0, 100.0] | < 0.0002 | 5.34 |

*Note: The "4-Fold Mean± SD" represents the average metric across the double-independent validation folds. Confidence intervals (CIs) were computed using 2000 participant-level bootstrap resamples. P-values and Cohen's d effect sizes were calculated using 5000 sequence-level permutations.*

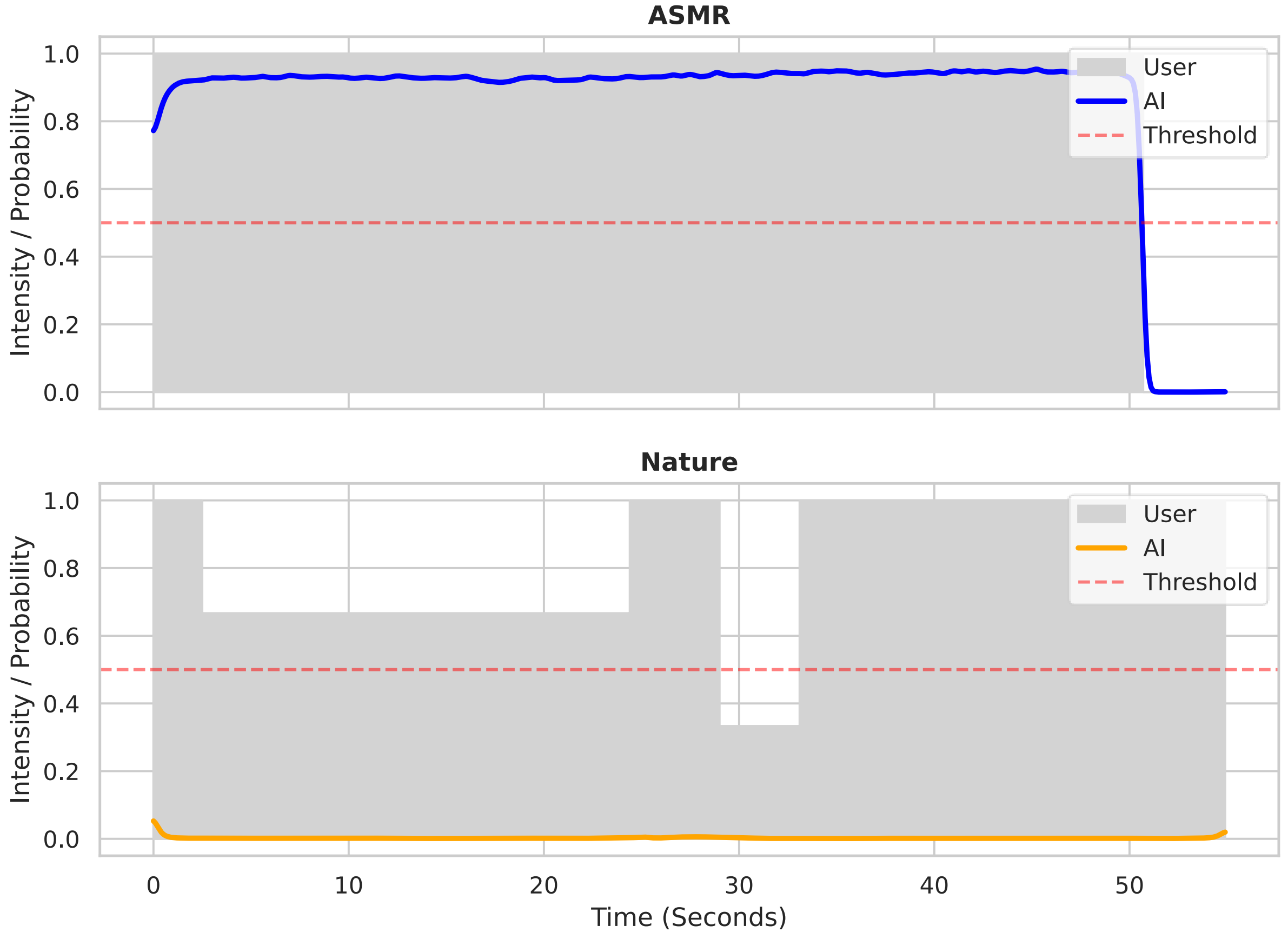

***Figure 6**. Temporal probability alignment of the multimodal fusion network against subjective ground truth (Subject 27). (Top) The ASMR condition demonstrates a high degree of temporal coupling, as the model's predicted tingle probability (blue line) closely follows the participant's binarized tingle report (grey shaded region), rising rapidly at onset and extinguishing at offset. (Bottom) The Nature control condition illustrates the model's high specificity; although the participant reported varying levels of subjective pleasantness (grey steps), the network correctly suppresses tingle prediction (orange line flatlining at 0.0), indicating that the learned features are separate from general positive feelings. The dashed red line represents the binary classification threshold (P=0.5).*

To qualitatively validate the temporal fidelity of the extracted features, a representative "Tinglegram" (**Figure 6**) was generated for a highly aligned test participant (Subject 27). In the ASMR condition (**Figure 6**, top panel), the predictions exhibit robust temporal phase-locking, with the probability curve (blue) rising in sync with the participant's reported tingle onset and sharply dropping when the sensation ended. This tight temporal connection highlights the precise hardware synchronization and the high temporal resolution of the continuous physiological data captured in the dataset. Importantly, the Nature control condition (**Figure 6**, bottom panel) highlights the discriminative power of these multimodal features. Although the participant reported fluctuating levels of subjective pleasantness during the control stimulus (indicated by the variable grey stepped blocks), the network correctly maintained a near-zero tingle probability throughout the trial. This confirms that the dataset successfully captures a specific, highly separable biological signature unique to the ASMR tingle state, rather than merely recording general positive affective valence or pleasantness.

### *Video-Level Classification Results*

The peak detection classification protocol was executed uniformly across the aggregated out-of-fold predictions of the double-independent 4-fold cross-validation framework to assess macro-level separability. Within the full multimodal fusion model, this thresholding methodology achieved perfect class separation, resulting in a mean video-level global accuracy of 100.0 percent (± 0.0) and a mean video-level macro F1-score of 100.0 percent (± 0.0). The model demonstrated a 95 percent confidence interval of [100.0, 100.0] and a statistically significant deviation from chance ($p < 0.0002$). Across the cumulative 68 fully held-out test sequences (comprising 34 ASMR trials and 34 Nature control trials), the model correctly identified the presence of ASMR triggers in every experimental trial while producing zero false positives across all control trials. This macro-level validation confirms that despite the high temporal variability of ASMR onset between participants, the synchronized multimodal features provide a robust, perfectly separable biological signature at the overall trial level.

### *Modality Ablation Study Results*

To isolate the contributions of the individual sensory streams and justify the multimodal nature of the dataset, the predictive performance of the full fusion model was compared against two unimodal ablations: a video-only spatial model and an audio-only acoustic model. A PPG-only model was purposefully excluded from this analysis because the cardiovascular efficacy of the PPG modality was already established in the previous sections. Moreover, from a physiological perspective, being a somatic effect of the ASMR affect, rather than being a causal trigger like audio or video, PPG should be used in conjunction with the primary triggering modalities to guide and biologically anchor the predictive architecture instead of being treated as a stand-alone feature to predict ASMR.

While both unimodal networks demonstrated predictive capability well above chance, most notably maintaining perfect Nature Specificity (100.0 ± 0.0), their overall performance was consistently superseded by the full fusion architecture. Specifically, removing physiological and acoustic features in the video-only baseline led to a drop in the mean global macro F1-score from 71.86 percent to 68.12 percent, alongside a drop in macro recall (78.53 percent). Similarly, the audio-only baseline achieved a mean macro F1-score of 69.94 percent. A paired bootstrap significance test (10,000 resamples) conducted across the 4-fold cross-validation results confirmed that the full multimodal fusion model provided a statistically significant improvement in the macro F1-score over both the video-only ($p = 0.0488$) and audio-only ($p = 0.0039$) baselines. These findings validate the dataset's core premise: while visual or acoustic contexts alone provide predictive signals for ASMR, their combination with synchronized physiological data delivers a more complete, complementary, and biologically grounded representation of the neurophysiological state associated with ASMR.

The comparative 4-fold cross-validation results across all three experimental configurations are detailed in Table 2, highlighting the superior predictive performance achieved through multimodal integration.

*Table 2 Comparative Mean 4-Fold Cross-Validation Results of Multimodal vs. Unimodal Architectures*

| EVALUATION METRIC (Global) | Video-Only Baseline (%) ± SD | Audio-Only Baseline (%) ± SD | Full Multimodal Fusion (%) ± SD |
|---|---|---|---|
| **Accuracy** | 71.99 ± 5.20 | 72.83 ± 6.33 | 75.51 ± 5.63 |
| **Macro F1** | 68.12 ± 6.34 | 69.94 ± 8.03 | 71.86 ± 8.86 |
| **Macro Precision** | 71.22 ± 4.77 | 71.87 ± 5.70 | 73.09 ± 8.42 |
| **Macro Recall** | 78.53 ± 5.76 | 80.81 ± 1.56 | 81.45 ± 6.34 |
| **Nature Specificity** | 100.0 ± 0.0 | 100.0 ± 0.0 | 100.0 ± 0.0 |
| **P-Value** | 0.0488 | 0.0039 | **(Ref)** |

*Note: Performance metrics are reported as percentages. Values represent the mean and standard deviation computed across the double-independent 4-fold cross-validation protocol. The full multimodal fusion model demonstrates a statistically significant improvement in the Macro F1-score over both unimodal baselines based on paired bootstrap resampling.*

### *Non-Temporal XGBoost Baseline Performance*

To assess the instantaneous predictive utility of the multimodal features independently of their temporal dynamics, the flattened dataset was evaluated using the XGBoost classifier. In the 4-fold cross-validation protocol, the non-temporal baseline attained a mean global accuracy of 70.86 ± 4.13 % and a mean global macro F1-score of 62.72 ± 10.21 %. The model achieved a macro precision of 68.22 ± 4.69 % and a macro recall of 72.98 ± 11.64 %. In line with the sequence models, the XGBoost classifier maintained near-perfect Nature Specificity at 99.99 ± 0.01 %, showing negligible false-positive prediction bias during baseline control stimuli.

While the tabular baseline demonstrates strong predictive capability well above chance, its overall performance shows a measurable degradation compared to the full sequence-to-sequence BiLSTM architecture (which achieved a 71.86% macro F1-score). This performance gap confirms the temporal nature of the ASMR response, validating that although immediate physiological and audiovisual states hold meaningful predictive signals, the ongoing temporal changes in these streams are necessary to fully capture the neurophysiological phenomenon.

## Limitations

While REST-ASMR provides a rigorously validated multimodal foundation for affective computing, certain limitations should be acknowledged. First, the participant group consists entirely of healthy university students (mean age 25.2 ± 4.6 years), so the physiological baselines and ASMR trigger sensitivities may not fully generalize to older adults or clinical populations (like those with severe generalized anxiety or insomnia). Future data collection efforts should aim to expand across broader age ranges and clinical groups to validate these models more widely. However, restricting this foundational dataset to a young, healthy cohort helped control for age-related cardiovascular comorbidities, medication-induced autonomic alterations, and confounding factors, thereby ensuring that the observed parasympathetic decelerations are clearly stimulus-driven.

Second, the subjective behavioral ground truth was captured utilizing a discrete 4-point Likert scale via keyboard input. While this effectively tracks the general onset and intensity of the sensation, it cannot capture the micro-level fluctuations of the subjective somatic experience. Future iterations of this experimental design could explore continuous input devices, such as rotary dials or pressure-sensitive joysticks, to capture higher-fidelity emotional gradients. Nevertheless, the implementation of a simple, discrete input mechanism was specifically chosen for this study to minimize cognitive and motor load, as highly demanding continuous reporting tasks can inadvertently disrupt the fragile ASMR state. The applied zero-order hold upsampling technique successfully bridged these discrete inputs into a dense 10 Hz time-series, enabling proper frame-by-frame modeling.

Finally, while the cohort size of 34 participants (yielding 272 full-length multimodal trials) provides sufficient statistical power for physiological validation, highly parameterized deep learning architectures typically benefit from large-scale data. Researchers utilizing REST-ASMR for complex deep learning applications are strongly advised to employ the strict, double-independent cross-validation frameworks detailed in this technical validation to prevent overfitting to individual physiological idiosyncrasies. Despite this, the dataset compensates with high temporal density and multimodality across audio, video, and PPG. Extracting features at 10 Hz yields 550 discrete time steps per trial, generating over 149,000 synchronized multimodal frames across the cohort. As demonstrated by our baseline experiments, this provides substantial sequential information for training Deep Learning architectures like BiLSTMs effectively.

## Data Availability

The REST-ASMR dataset presented and analyzed in this paper is available in the Zenodo repository[29] at https://doi.org/10.5281/zenodo.18881334.

## Code Availability

The code to replicate the experiments described in this paper is released on GitHub: https://github.com/Tushardas2663/REST-ASMR-Pipeline and is permanently archived on Zenodo[35] at https://doi.org/10.5281/zenodo.18881306. The repository is orchestrated by main.py, which executes the backend pipeline encompassing log parsing, multimodal feature extraction, and the 4-fold cross-validation training for the sequence-to-sequence and XGBoost models. A front-end Jupyter Notebook (REST_ASMR_Analysis.ipynb) is provided to reproduce the statistical validations, performance tables, and data visualizations reported in the manuscript. The repository is structurally divided into a /pipeline directory for Python modules and an empty /data directory where users must deposit the raw Zenodo dataset files (/log, /ppg, /stim) prior to execution. Finally, to ensure strict reproducibility despite the inherent non-determinism of GPU-accelerated atomic operations during backpropagation, the exact serialized inference artifacts (paper_fusion_results_4_fold.pkl) generated during our technical validation are included within the repository.

## Author Contributions

T.D. (Conceptualization, Methodology, Software, Validation, Formal analysis, Data curation, Visualization, Writing—original draft, Writing—review & editing), D.H. (Conceptualization, Investigation, Data curation, Methodology, Writing—review & editing), K.K.S. (Supervision, Resources, Formal analysis, Writing—review & editing), and H.M.K. (Conceptualization, Resources,

Data curation, Investigation, Funding acquisition, Project administration, Supervision, Writing—review & editing). All authors reviewed the manuscript.

# Competing Interests

The authors declare no competing interests.

# Funding

This study was supported by the Japan Society for the Promotion of Science (JSPS) KAKENHI grant 22K18659.

# Figure Legends

- Figure 1. Experimental protocol timeline and counterbalanced block design. The 12-minute data acquisition session utilized a repeated-measures design to alternate the presentation order of the stimuli. To mitigate potential order and carryover effects, the 34 participants were counterbalanced into two groups. Sequence 1 (top) presented the ASMR stimuli block first, while Sequence 2 (bottom) presented the non-social Nature control stimuli block first. Each session commenced with a 60-second baseline resting period, followed by the first 240-second stimulus block (comprising four category-specific 60-second videos presented in a pseudo-randomized order). A 120-second washout resting period separated the two stimulus blocks to allow cardiovascular metrics to return to baseline. Following the presentation of the second 240-second stimulus block, a final 60-second resting period concluded the acquisition session.

- Figure 2. Multimodal data processing and temporal alignment pipeline. The parallel processing streams used to synchronize various data modalities into a single, dense matrix at a consistent 10 Hz temporal resolution are depicted in the schematic. In order to extract spatial (ResNet-18) and acoustic (MFCCs and spectral) features, the audiovisual stream (left) discretizes raw stimuli. The physiological stream (center) uses video-level Z-score normalization and Fourier-based downsampling to separate the raw 2,000 Hz photoplethysmography (PPG) signal. The behavioral stream (right) creates a dense continuous label by parsing sparse subjective rating logs, applying the required latency corrections, enforcing a 5-second warm-up buffer, and using zero-order hold upsampling. A final standardized data matrix with 550 time steps by 545 dimensions (544 predictive features and 1 binary state label) is created by the convergence of these three independent streams for each participant-video trial.

- Figure 3. Group-level temporal dynamics of continuous behavioral annotations. Mean time-series plots for the two highest-performing ASMR stimuli, (A) Video 2 and (B) Video 3, visualize the temporal consensus across all active participants. The shaded regions represent the standard error of the mean (SEM). The contrasting temporal architectures, ranging from multi-phasic, episodic triggering in Video 2 to steady, accumulative physiological build-up in Video 3, highlight the diverse temporal dynamics captured within the dataset, together with synchronized Inter-Subject Agreement across different trigger types.

- Figure 4. Participant-level cardiovascular correlation across experimental conditions. Boxplots illustrating the distribution of Spearman rank correlation coefficients (r) between instantaneous pulse rate (PR) and subjective intensity ratings for the ASMR and Nature control stimuli. The ASMR condition exhibits a negative mean correlation, reflecting the parasympathetic cardiovascular deceleration accompanying the ASMR tingle. Conversely, the Nature condition trends positively, consistent with mild visual engagement and baseline physiological arousal. Individual participant data points are overlaid to demonstrate the population variance. The group-level difference is statistically significant ($p = 0.0454$) based on a paired t-test of the Fisher Z-transformed coefficients.
- Figure 5. Multimodal deep learning architecture for frame-by-frame ASMR state prediction. Raw audiovisual and physiological streams are synchronized and sampled at 10 Hz. For the visual modality (top), a pre-trained ResNet-18 network extracts 512-dimensional spatial features from individual video frames. The acoustic modality (middle) extracts a 31-dimensional feature vector comprising Mel-Frequency Cepstral Coefficients (MFCCs) and spectral properties. The physiological modality (bottom) utilizes Z-score normalized continuous photoplethysmography (PPG) signals to yield a 1-dimensional feature. These discrete streams are horizontally concatenated to form a unified 544-dimensional multimodal feature vector for each 100-millisecond time step. A two-layer Bidirectional Long Short-Term Memory (BiLSTM) network, with a hidden dimension of 128 and a dropout rate of 0.3, processes this dense sequential representation to model complex temporal dependencies. A fully connected classification layer with a sigmoid activation function then outputs the frame-level binary prediction, classifying the participant's state as either experiencing a tingle (1) or not (0).

- Figure 6. Temporal probability alignment of the multimodal fusion network against subjective ground truth (Subject 27). (Top) The ASMR condition demonstrates a high degree of temporal coupling, as the model's predicted tingle probability (blue line) closely follows the participant's binarized tingle report (grey shaded region), rising rapidly at onset and extinguishing at offset. (Bottom) The Nature control condition illustrates the model's high specificity; although the participant reported varying levels of subjective pleasantness (grey steps), the network correctly suppresses tingle prediction (orange line flatlining at 0.0), indicating that the learned features are separate from general positive feelings. The dashed red line represents the binary classification threshold ($P=0.5$).